\documentclass[aoas,nameyear,dvips]{arximspdf}
\usepackage{graphics}

\doi{10.1214/09-AOAS311}
\volume{4}
\issue{2}
\pubyear{2010}
\firstpage{535}
\lastpage{566}

\makeatletter
\newcommand{\parsesicm}[1]{#1}
\newcommand{\parpenkicm}[1]{#1}
\newcommand{\Ev}{E} 
\newcommand{\Av}{\mathbf{A}}
\newcommand{\zv}{\mathbf{z}}
\newcommand{\D}{}
\newcommand{\zitj}[1][]{\vec{z}_{#1 i \rightarrow j}} 
\newcommand{\zjfi}[1][]{\vec{z}_{#1 j \leftarrow i}} 
\newcommand{\zitodot}{\vec{z}_{i \rightarrow\cdot}} 
\newcommand{\zifromdot}{\vec{z}_{i \leftarrow\cdot}} 
\newcommand{\zitjs}[2][]{z_{#1 i \rightarrow j, #2}} 
\newcommand{\zjfis}[2][]{z_{#1 j \leftarrow i, #2}} 
\newcommand{\gammai}{\vec\gamma_{i}}
\newcommand{\gammadot}{\vec\gamma_{\cdot}}
\newcommand{\zdotTodot}{\vec{z}_{\cdot\rightarrow\cdot}}
\newcommand{\zdotFromdot}{\vec{z}_{\cdot\leftarrow\cdot}}
\newcommand{\gammaj}{\vec\gamma_{j}}
\newcommand{\eij}{e_{ij}}
\newcommand{\Buv}{\beta_{u,v}}
\newcommand{\Ezitju}{{\langle z_{i\rightarrow j,u}\rangle}_{q_z}}
\newcommand{\Ezjfiv}{{\langle z_{j\leftarrow i,v}\rangle}_{q_z}}
\newcommand{\Estz}[1]{\langle{#1}\rangle_{q_z}}
\newcommand{\Estgamma}[1]{\langle{#1}\rangle_{q_\gamma}}
\newcommand{\ziFromjSubk}{z_{i \leftarrow j, k}}
\newcommand{\tildegamma}{\tilde\gamma}
\newcommand{\tildeSigma}{\tilde\Sigma}

\newcommand{\Bkl}{\beta_{k,l}}
\newcommand{\hatBkl}{\hat{\beta}_{k,l}}
\newcommand{\hatgamma}{\hat\gamma}
\newcommand{\KFx}{\hat{\mu}}
\newcommand{\KFP}{P}
\makeatother

\begin{document}
\begin{frontmatter}

\title{A state-space mixed membership blockmodel for dynamic network tomography}
\runtitle{Dynamic network tomography}

\begin{aug}
\author[a]{\fnms{Eric P.} \snm{Xing}\corref{}\thanksref{tf}\ead[label=e1]{epxing@cs.cmu.edu}},
\author[a]{\fnms{Wenjie} \snm{Fu}\ead[label=e2]{wenjief@cs.cmu.edu}} \and
\author[a]{\fnms{Le} \snm{Song}\ead[label=e3]{lesong@cs.cmu.edu}}
\thankstext{tf}{Supported by Grant ONR N000140910758, NSF DBI-0640543, NSF DBI-0546594, NSF
IIS-0713379, DARPA CS Futures II award, and an Alfred P. Sloan Research
Fellowship.}
\runauthor{E.~P. Xing, W. Fu and L. Song}
\affiliation{Carnegie Mellon University}
\address[a]{E. P. Xing\\
W. Fu\\
L. Song\\
School of Computer Science
\\Carnegie Mellon University\\
5000 Forbes Ave, Wean Hall 4212\\
Pittsburgh, Pennsylvania 15213\\
USA\\
\printead{e1}\\
\phantom{E-mail:\ }\printead*{e2}\\
\phantom{E-mail:\ }\printead*{e3}} 
\end{aug}

\received{\smonth{1} \syear{2009}}
\revised{\smonth{11} \syear{2009}}

%
\begin{abstract}
In a dynamic social or biological environment, the interactions
between the actors can undergo large and systematic changes.
In this paper we propose a model-based approach to analyze what we will
refer to as
the \textit{dynamic tomography} of such time-evolving networks. Our approach
offers an intuitive but powerful tool to infer the semantic underpinnings
of each actor, such as its social roles or biological functions, underlying
the observed network topologies.
Our model builds on earlier work on a mixed membership stochastic blockmodel
for static networks, and the state-space model for tracking object trajectory.
It overcomes a major limitation of many current network inference techniques,
which assume that each actor plays a unique and invariant role that accounts
for all its interactions with other actors; instead, our method models the
role of each actor as a time-evolving \textit{mixed membership vector} that
allows actors to behave differently over time and carry out different
roles/functions when interacting with different peers, which is closer
to reality. We present an efficient algorithm for approximate inference
and learning using our model; and we applied our model to analyze
a social network between monks (i.e., the Sampson's network), a dynamic
email communication network between the Enron employees, and a rewiring
gene interaction network of fruit fly collected during its full life cycle.
In all cases, our model reveals interesting patterns of the
dynamic roles of the actors.
\end{abstract}

%

\end{frontmatter}

\section{Introduction}

Networks are a fundamental form of representation of complex systems.
In many problems arising in biology, social sciences, and various other
fields, it is often necessary to analyze populations of entities such
as molecules or individuals, also known as ``actors'' in some network
literature, interconnected by a set of relationships such as regulatory
interactions, friendships, and communications. Studying networks of
these kinds can reveal a wide range of information, such as how
molecules/individuals organize themselves into groups, which molecules
are the key regulator or which individuals are in positions of power,
and how the patterns of biological regulations or social interactions
are likely to evolve over time.

In this paper we investigate an intriguing statistical inference
problem of interpreting the dynamic behavior of temporally evolving
networks based on a concept known as \textit{network tomography}. Borrowed
from the vocabulary of magnetic resonance imaging, the term ``network
tomography'' was first introduced by~\citet{Vardi96} to refer to the
study of a network's internal characteristics using information derived
from the observed network.
In most real-world complex systems such as a social network or a gene
regulation network, the measurable attributes and relationships of
vertices (or nodes) in a network are often functions of latent temporal
processes of events which can fluctuate, evolve, emerge, and terminate
stochastically. Here we define \textit{network tomography} more
specifically as the latent semantic underpinnings of entities in both
static and dynamic networks.
For example, it can stand for the latent class labels, social roles, or
biological functions undertaken by the nodal entities, or the measures
on the affinity, compatibility, and cooperativity between nodal states
that determine the edge probability. Our goal is to develop a
statistical model and algorithms with which such information can be
inferred from dynamically evolving networks via posterior inference.

We will concern ourselves with three specific real world time-evolving
networks in our empirical analysis: (1) the well-known Sampson's
undirected social networks [\citet{SampsonMonk}] of 18 monks over 3 time
episodes, which are recorded during an interesting timeframe that
preludes a major conflict followed by a mass departure of the monks,
and therefore an interesting example case to infer nodal causes behind
dramatic social changes; (2) the time series of email-communication
networks of ENRON employees before and during the collapse of the
company, which may have recorded interesting and perhaps sociologically
illuminating behavioral patterns and trends under various business
operation conditions; and (3) the sequence of gene interaction networks
estimated at 22 time points during the life span of \textit{Drosophila
melanogaster}, a fruit fly commonly used as a lab model to study the
mechanisms of animal embryo development, which captures transient
regulatory events such as the animal aging.

Inference of network tomography is fundamental for
understanding the organization and function of complex relational
structures in natural, sociocultural, and technological systems such as
the ones mentioned above. In a social system such as a company employee
network, network tomography can
capture the latent social roles of individuals; inferring such roles
based on the social interactions among individuals is fundamental
for understanding the importance of members in a network, for
interpreting the social structure of various communities in a
network, and for modeling the behavioral, sociological, and
even epidemiological processes mediated by the vertices in a network. In
systems biology, network tomography often translates to latent
biochemical or genetic functions of interacting molecules
such as proteins, mRNAs, or metabolites in a regulatory circuity;
elucidating such functions based on the topology of molecular
networks can advance our understanding of the mechanisms of how a
complex biological system regulates itself and reacts to stimuli.
More broadly, network tomography can lead to important insights to
the robustness of network structures and their vulnerabilities, the
cause and consequence of information diffusion, and the mechanism of
hierarchy and organization formation. By appropriately modeling
network tomography, a network analyzer can also simulate and reason
about the generative mechanisms of networks, and discover changing
roles among actors in networks, which will be relevant for activity and anomaly
detection.

There has been a variety of successes in network analysis based on
various formalisms. For example, researchers have found trends in a
wide variety of large-scale networks, including scale-free and
small-world properties [\citet{BaraAlbe1999}; \citet{Klei2000}]. Other
successes include the formal characterization of otherwise intuitive
notions, such as ``groupness'' which can be formally characterized in
the networks perspective using measures of structural cohesiveness and
embeddedness [\citet{moody03}], detecting outbreaks~[\citet{Leskovec07kdd}], and characterizing macroscopic properties of various
large social and information networks~[\citet{Leskovec08www}].
Additionally, there has been progress in statistical modeling of social
networks, traditionally focusing on descriptive models such as the
exponential random graph models, and more recently moving toward
various latent space models that estimate an embedding of the network
in a latent semantic space, as we review shortly in Section~\ref{review}.
A major limitation of most current methods for network modeling and
inference~[\citet{HoffRaftHand2002}; \citet{McCallumICML06}; \citet{handcockRSS06}]
is that they assume each actor, such as a social
individual or a biological molecule in a network, undertakes a
single and invariant role (or functionality, class label, etc.,
depending on the domain of interest), when interacting with other
actors. In many realistic social and biological scenarios, every
actor can play multiple roles (or under multiple influences) and the
specific role being played depends on whom the actor is
interacting with; and the roles undertaken by an actor can change
over time. For example, during a developmental process or an immune
response in a biological system, there may exist multiple underlying
``themes'' that
determine the functionalities of each molecule and their
relationships to each other, and such themes are dynamical and
stochastic. As a result, the molecular networks at each time point
are context-dependent and can undergo systematic rewiring, rather
than being i.i.d. samples from a single underlying distribution, as
assumed in most current biological network studies. We are
interested in understanding the mechanisms that drive the temporal
rewiring of biological networks during various cellular and
physiological processes, and similar phenomena in time-varying
social networks.

In this paper we propose a new Bayesian approach for
network tomographic inference that will capture the multi-facet,
context-specific, and temporal nature of an actor's role in large,
heterogeneous, and evolving dynamic networks. The proposed method
will build on a modified version of the \textit{mixed membership
stochastic blockmodel} (MMSB)~[\citet{AiroBleiFienXing2008}], which
enables network links
to be realized by role-specific \textit{local} connection
mechanisms; each link is underlined by a separately chosen latent
functional cause, and each vertex can have fractional involvement in
multiple functions or roles which are captured by a \textit{mixed
membership vector}, thereby the proposed model supports analyzing
patterns of interactions between actors via statistically inferring an
``embedding'' of a network in a latent ``tomographic-space'' via the
mixed membership vectors. For example, the characteristics of group
profiles of actors revealed by the mixed membership vectors can offer
important and intuitive community structures in the networks in question.


Modeling embedding of networks in latent state space offers an
intuitive but powerful approach to infer the semantic underpinnings
of each actor, such as its biological or social roles or other
entity functions, underlying the observed network topologies. Via
such a model, one can map every actor in a network to a position in
a low-dimensional simplex, where the roles/functions of the actors are
reflected in the role- or functional-\textit{coordinates} of the actors
in the latent space and the relationships among actors are reflected
in their Euclidian distances.
We can naturally capture the dynamics of role evolution of actors in
such a tomographic-space, and other latent dynamic processes driving
the network evolution by furthermore applying a state-space model
(SSM) popular in object tracking over the positions of the
tomographic-embeddings of all actors, where a \textit{logistic-normal}
mixed membership stochastic blockmodel is employed as the
emission model to define time-specific condition likelihood of the
observed networks over time. The resulting model shall be formally
known as a \textit{state-space mixed membership stochastic blockmodel},
but, for simplicity, in this paper we will refer to it as a \textit{dynamic MMSB} (or, in short, dMMSB); and we will show that this model
allows one to infer the trajectory of the roles of each actor based
on the posterior distribution of its role-vector.

Given network data, the dMMSB can be learned based on the maximum
likelihood principle using a variational EM
algorithm~[\citet{zoubin1}; \citet{xingGMF}; \citet{xingLoNTAM}], the resulting network
parameters reveal
not only mixed membership information of each actor over time, but
also other interesting regularities in the network topology. We will
illustrate this model on the well-known Sampson's monk social network,
and then apply it to the time series of email network from Enron and
the sequence of time-varying genetic interaction networks
estimated from the Drosophila genome-wise microarray time series,
and we will present some previously unnoticed dynamic behaviors of
network actors in these data.

The remaining part of the paper is organized as follows. In Section~\ref{review} we briefly review some related work.
In Section~\ref{sec3} we present the dMMSB model in detail. A~Laplace variational EM
algorithm for approximate inference under dMMSB will be described in
Section~\ref{sec:VI}. In Section~\ref{sec5} we present case studies on the monks
network, the Enron network, and the Drosophila gene network using
dMMSB, along with
some simulation based validation of the model. Some discussions will
be given in Section~\ref{sec6}. Algebraic details of the derivations of the
inference algorithm are provided in the \hyperref[appendix]{Appendix}.

\section{Related work}\label{review}

There is a vast and growing body of literature on model-based statistical
analysis of network data, traditionally focusing on descriptive
models such as the exponential random graph models (ERGMs) [\citet{FranStra1986}; \citet{WassPatt1996}], and more recently
moving toward more generative types of models such as those that
model the network structure as being caused by the actors' positions
in a latent ``social space'' [\citet{HoffRaftHand2002}].
Among these models, some variants of the ERGMs, such as the \textit{stochastic block models}~[\citet{HollLaskLein1983}; \citet{FienMeyeWass1985}; \citet{WassPatt1996}; \citet{Snij2002}],
cluster network vertices based on their structural
equivalency~[\citet{LorrWhit1971}]. The latent space models (LSM)
instead project nodes onto a latent space, where their similarities
can be visualized and explored~[\citet{HoffRaftHand2002}; \citet{Hoff2003b}; \citet{handcockRSS06}]. The mixed
membership stochastic blockmodel proposed in~\citeauthor{xingLinkKDD05} (\citeyear{xingLinkKDD05,AiroBleiFienXing2008}) integrates ideas
from these models, but went further by allowing each node to belong
to multiple blocks (i.e., groups) with fractional membership.
Variants of the mixed membership model have appeared in population
genetics~[\citet{PritStepDonn2000}], text
modeling~[\citet{BleiJordNg2003}], analysis of multiple disability
measures~[\citet{ErosFien2005}], etc. In most of these cases mixed
membership models are used as a latent-space projection method to
project high-dimensional attribute data into a lower-dimensional
``aspect-space,'' as a normalized \textit{mixed membership vector},
which reflects the weight of each latent aspect (e.g., roles,
functions, topics, etc.) associated with an object~[\citet{ErosFienLaff2004}]. The mixed
membership vectors often serve as a surrogate of the original data
for subsequent analysis such as classification~[\citet{BleiNgJord2003}].
The MMSB model developed earlier has been applied for
role identification in Sampson's 18-monk social
network and functional prediction in a protein--protein interaction
network (PPI)~[\citeauthor{xingLinkKDD05} (\citeyear{xingLinkKDD05,AiroBleiFienXing2008})].
It uses the aforementioned mixed membership vector to define an
actor-specific multinomial distribution, from which specific actor
roles can be sampled when interacting with other actors. For each
monk, it yields a multi-class social-identity prediction which
captures the fact that his interactions with different other monks
may be under different social contexts. For each protein, it yields a
multi-class functional prediction which captures the fact that its
interactions with different proteins may be under different functional
contexts.

We intend to use the state-space model (SSM) popular in object tracking
and trajectory modeling for inferring underlying functional changes
in network entities, and sensing emergence and
termination of ``function themes'' underlying network sequences. This
scheme has been adopted in a number of recent works on extracting
evolving topical themes in text documents~[\citet{bleiDTM}; \citet{McCallumKDD06}] or author
embeddings~[\citet{SarkarNIPS06}] based on author, text, and reference
networks of archived publications.

\section{Modeling dynamic network tomography}\label{sec3}

Consider a temporal series of networks $\{{G}^{(1)}, \ldots, {G}^{(T)}\}$
over a vertex set $V$, where ${G}^{(t)} \equiv\{V, E^{(t)}\}$ represents
the network observed at time $t$. In this paper we assume that
$N=|V|$ is invariant over time; thus, ${E}^{(t)} \equiv\{e^{\D
(t)}_{i,j}\}_{\D i,j=1}^{\D N}$ denote the set of (possibly
transient)\vspace*{1pt} links at time $t$ between a fixed set of $N$
vertices.

To model both the multi-class nature of every vertex in a network
and the latent semantic characteristics of the vertex-classes and
their relationships to inter-vertices interactions, we assume that
at any time point, every vertex $v_i \in V$ in the network, such as
a social actor or a biological molecule, can undertake multiple
roles or functions realized from a predefined latent tomographic
space according to a time-varying distribution $P_t(\cdot)$; and the
weights (i.e., proportion of ``contribution'') of the involved
roles/functions can be represented by a normalized vector
$\vec{\pi}^{\D(t)}_i$ of fixed dimension $K$. We refer to each role,
function, or other domain-specific semantics underlying the vertices
as a \textit{membership} of a latent class. Earlier stochastic
blockmodels of networks restricted each vertex to belong to
a single and invariant membership.
In this paper we assume that each vertex can have \textit{mixed memberships},
that is, it can undertake multiple roles/functions within a single
network when interacting with various network neighbors with
different roles/functions, and the vector of proportions of the
mixed-memberships, $\vec{\pi}^{\D(t)}_i$, can evolve over time. Furthermore,
we assume that the links between vertices are instantiated
stochastically according to a \textit{compatibility function} over the
roles undertaken by the vertex-pair in question, and we define the
compatibility coefficients between all possible pair of roles using
a time-evolving role-compatibility matrix ${B}^{(t)} \equiv\{\beta^{\D(t)}_{k,l}\}$.

\subsection{Static mixed membership stochastic blockmodel}\label{sec:sMMSB}

Under a basic MMSB model, as first proposed
in~\citet{xingLinkKDD05}, network links can be realized by a
role-specific \textit{local} interaction mechanism: the link between
each pair of actors, say, $(i,j)$, is instantiated according to the
latent role specifically undertaken by actor~$i$ when it is to
interact with $j$, and also the latent role of $j$ when it is to
interact with $i$. More specifically, suppose that each different
role-pair, say, roles $k$ and~$l$, between actors has a unique
probability distribution $P(\cdot|\beta_{k,l})$ of having a link
between actor pairs with that role combination, then a basic mixed
membership stochastic blockmodel posits the following generative
scheme for a static network:
%
%
\begin{enumerate}
\item For each vertex $i$, draw the mixed-membership vector:
$\vec{\pi}_i \sim P(\cdot|\theta)$.
\item For each possible interacting vertex $j$ of vertex $i$, draw the
link indicator
$e_{\D i,j} \in\{0,1\}$ as follows:
\begin{itemize}
\item draw latent roles $\zitj[\D] \sim\operatorname{Multinomial}(\cdot|\vec{\pi
}_i, 1)$,
$\zjfi[\D] \sim\operatorname{Multinomial}(\cdot|\vec{\pi}_j, 1)$,
where $\zitj[\D]$ denotes the role of actor $i$ when it is to interact
with $j$,
and $\zjfi[\D]$ denotes the role of actor $j$ when it is approached by
$i$. Here
$\zitj[\D]$ and $\zjfi[\D]$ are unit indictor vectors in which one
element is one and the rest are zero; it represents the $k$th role if
and only if the $k$th element of the vector is one, for example, $\zitjs
[\D]{k}=1$ or $\zjfis[\D]{k}=1$
\item and draw $e_{\D i,j}  | (\zitjs[\D]{k}=1,
\zjfis[\D]{l}=1) \sim \operatorname{Bernoulli}(\cdot|{\beta}_{\D k,l})$.
\end{itemize}
\end{enumerate}
%

Specifically, the generative model above defines a conditional
probability distribution of the relations $\Ev=\{e_{i,j}\}$ among
vertices in a way that reflects naturally interpretable latent
semantics (e.g., roles, functions, cluster identities) of the
vertices. The link $e_{i,j}$ represents a binary actor-to-actor
relationship. For example, the existence of a link could mean that a
package has been sent from one person to another, or one has a
positive impression on another, or one gene is regulated by another.
Each vertex $v_i$ is associated with a set of latent membership
labels $\{\zitodot, \zifromdot\}$ (if
the links are undirected, as in a PPI, then we can ignore the
asymmetry of ``$\rightarrow$'' and ``$\leftarrow$''). Thus, the semantic
underpinning of each interaction between vertices is captured by a
pair of instantiated memberships unique to this interaction; and the
nature and strength of the interaction is controlled by the
compatibility function determined by this pair of memberships'
instantiation. For example, if actors A and C are of role $X$ while
actors B and D are of role $Y$, we may expect that the relationship
from A to B is likely to be the same as relationship from C to D,
because both of them are from a role-$X$ actor to a role-$Y$ actor.
In this sense, a role is like a class label in a classification
task. However, under an MMSB model, an actor can have
different role instantiations when interacting with different
neighbors in the same network.

The role-compatibility matrix ${B} \equiv\{\beta_{k,l}\}$
decides the affinity between roles. In some cases, the diagonal
elements of the matrix may dominate over other elements, which means
actors of the same role are more likely to connect to each other.
In the case where we need to model differential preference among
different roles, richer block patterns can be encoded in the
role-compatibility matrix. The flexibility of the choices of the $B$
matrix give rise to strong expressivity of the model to deal with
complex relational patterns. If necessary, a prior distribution over
elements in $B$ can be introduced, which can offer desirable
smoothing or regularization effects.

Crucial to our goal of role-prediction and role-evolution modeling
for network data is the so-called mixed membership vector ${\vec\pi
}_i$, also referred to as ``role vector,'' of
the \textit{mixed-membership coefficients} in the above generative
model, which represents the overall role spectrum of each actor and
succinctly captures the probabilities of an actor involving in
different roles when this actor interacts with another actor. Much of
the expressiveness
of the mixed-membership models lies in the choice of the prior
distribution for the mixed-membership coefficients ${\vec\pi}_i$,
and the prior for the interaction coefficients $\{\beta_{k,l}\}$.
For example, in~\citeauthor{xingLinkKDD05} (\citeyear{xingLinkKDD05,AiroBleiFienXing2008}), a
simple Dirichlet prior was
employed because it is conjugate to the multinomial distribution
over every latent membership label $\{\zitodot, \zifromdot\}$ defined
by the relevant $\vec\pi_i$. In this paper, to
capture nontrivial correlations among the weights (i.e., the
individual elements within ${\vec\pi}_i$) of all latent roles of a
vertex, and to allow one to introduce dynamics to the roles of each
actor when modeling temporal processes such as a cell cycle, we
employ a logistic-normal distribution over a
simplex~[\citet{AitchisonShen80}; \citet{Aitchison}; \citet{xingLoNTAM}]. The resulting
model is referred to as a logistic-normal MMSB, or simply LNMMSB.

Under a logistic normal prior, assuming a centered logistic transformation,
the first sampling step for
$\vec{\pi}_i \equiv[\pi_{i,1}, \ldots, \pi_{i,K}]$ in the canonical mixed
membership generative model above can be broken down into
two sub-steps: first draw $\vec{\gamma}_i$ according to
%
\begin{equation}
\vec{\gamma}_i \sim\operatorname{Normal}(\vec{\mu}, \Sigma);
\end{equation}
then map it to the simplex via the following \textit{logistic}
transformation:
%
\begin{eqnarray}
\pi_{\D i,k} = \operatorname{exp}\{\gamma_{\D i,k} - C({\vec\gamma}_{\D i})\}
\qquad  \forall k=1,\ldots,K,
\end{eqnarray}
where
\begin{eqnarray}\label{logPartitionFunction}
C({\vec\gamma}_{\D i})=\log \Biggl(\sum_{k=1}^K \operatorname{exp}\{\gamma_{\D i,k}\} \Biggr).
\end{eqnarray}
Here $C({\vec\gamma}_{\D i})$ is a normalization constant (i.e., the
log partition function). Due to the normalizability constrain of the
multinomial parameters, $\vec{\pi}_i$ only has $K-1$ degree of
freedom. Thus, we only need to draw the first $K-1$ components of
$\vec{\gamma}_{\D i}$ from a $(K-1)$-dimensional multivariate
Gaussian, and leave $\gamma_{\D i,K}=0$. For simplicity, we omit
this technicality in the forthcoming general description and operation
of our
model.

Under a dynamic network tomography model, the prior distributions of
role weights of every vertex $P_t(\cdot),$ and the role-compatibility
matrix $B$, can both evolve over time. Conditioning on the observed
network sequence $\{{G}^{(1)}, \ldots, {G}^{(T)}\}$, our goal is to infer
the trajectories of role vectors $\vec{\pi}^{\D(t)}_i$ in the
latent social space or biological function space. In the following,
we present a generative model built on elements from the classical state-space
model for linear dynamic systems and the static logistic normal MMSB
described above for random graphs for this purpose.

\subsection{Dynamic logistic-normal mixed membership stochastic blockmodel}\label{sec:dMMSB}

We propose to capture the dynamics of network evolution
at the level of both the prior distributions of the mixed membership
vectors of vertices, and the compatibility functions governing
role-to-role relationships.
In this way we capture the dynamic behavior of the generative system of
both vertices and
relations. Our basic model structure is based on the well-known
state-space model, which defines a linear dynamic transformation of the mixed
membership priors over adjacent time points:
%
\begin{equation}\label{eq:ssm}
\vec{\mu}^{\D(t)} = \Av\vec{\mu}^{\D(t-1)} + \vec{w}^{\D(t)}\qquad \mbox{for } t \geq1,
\end{equation}
where $\vec{\mu}^{\D(t)}$ represents the mean parameter of the
prior distribution of the transformed mixed membership vectors of all vertices
at time $t$, and $\vec{w}^{\D(t)}\sim\mathcal{N}(0,\Phi) $ represents
normal transition noise for the mixed membership prior, and the
transition matrix $\Av$ shapes the trajectory of temporal
transformation of the prior.
The LNMMSB model defined above now functions as an emission
model within the SSM that defines the conditional likelihood of the
network at each time point. Note that the linear system on
$\vec{\mu}^{\D(t)}$ can lead to a bursty dynamics for latent admixing
vector $\pi^{\D(t)}_i$ through the LNMMSB emission model.
Starting from this basic structure, we propose to develop a
dynamical model for tracking underlying functional changes in
network entities and sensing emergence and termination of ``function
themes.''

\begin{figure}

\includegraphics{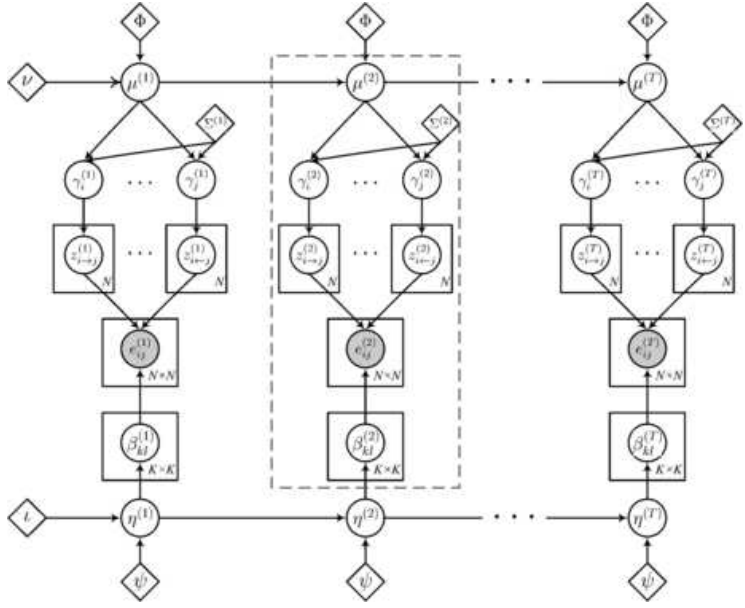}

\caption{A graphical model representation of the dynamic
logistic-normal mixed membership stochastic blockmodel. The part
enclosed by the dotted lines is a logistic-normal MMSB.} \label{fig0}
\end{figure}

Given a sequence of network topologies over the same set of nodes,
here is an outline of the generative process under such a model (a
graphical model representation of this model is illustrated in
Figure~\ref{fig0}):
\begin{itemize}
\item \textit{State-space model for mixed membership prior}:
\begin{itemize}
\item
$\vec{\mu}^{(1)} \sim\operatorname{Normal}(\nu,\Phi)$,
\parsesicm{sample the mean of the mixed membership prior at time 1.}
\end{itemize}
For $t=1, \ldots, T$:
\begin{itemize}
\item
$\vec{\mu}^{(t)} = \operatorname{Normal}(\Av\vec{\mu}^{(t-1)},\Phi)$,
\parsesicm{sample the means of the mixed membership
priors over time.}
\end{itemize}
\item \textit{State-space model for role-compatibility matrix}:
\item[\phantom{$\bullet$}]For $k=1,\ldots,K$ and $k'=1,\ldots,K$,
\begin{itemize}
\item
${\eta}_{k,k'}^{\D(1)} \sim\operatorname{Normal}(\iota, \psi)$,
\parsesicm{sample the compatibility coefficient between role $k$ and
$k'$ at time 1.}
\end{itemize}
For $t=1, \ldots, T$:
\begin{itemize}
\item
${\eta}_{k,k'}^{\D(t)} \sim\operatorname{Normal}(b \eta_{k,k'}^{\D
(t-1)},\psi)$,  \parsesicm{sample compatibility coefficients over subsequent time points.}
\item
$\beta_{k,k'}^{\D(t)}={\exp(\eta_{k,k'}^{\D(t)})}/{(\exp(\eta_{k,k'}^{\D(t)})+1)}$,
\parsesicm{compute compatibility probabilities via logistic transformation.}
\end{itemize}
\item\textit{Logistic-normal mixture membership model for networks:}
\item[\phantom{$\bullet$}]For each node $n = 1, \ldots, N$, at each time point $t=1, \ldots, T$:
\begin{itemize}
\item $\vec{\pi}^{\D(t)}_i \sim\operatorname{LogisticNormal}( \vec{\mu
}^{(t)}, \Sigma^{(t)} )$,
\parpenkicm{sample a $k$ dimensional \textit{mixed membership} vector.}
\end{itemize}
For each pair of nodes $(i,j) \in[1,N] \times[1,N]$:
\begin{itemize}
\item${\zitj}^{\D(t)}  \sim\operatorname{Multinomial}(\vec{\pi}^{\D
(t)}_i, 1 )$,
\parpenkicm{sample membership indicator for the donor,}
\item${\zjfi}^{\D(t)}  \sim\operatorname{Multinomial}(\vec{\pi}^{\D
(t)}_j, 1 )$,
\parpenkicm{sample membership indicator for the acceptor,}
\item
$e^{\D(t)}_{i,j} \sim\operatorname{Bernoulli}({\zitj[\D]}^{\D(t)\prime}B^{\D(t)}
{\zjfi[\D]}^{\D(t)})$,
\parpenkicm{sample the links between nodes.}
\end{itemize}
\end{itemize}
%

Specifically, we assume that the mixed membership vector
$\vec{\pi}$ for each actor follows a time-specific logistic
normal prior $\mathcal{LN}({\vec\mu}^{\D(t)},\Sigma^{\D(t)})$, whose
mean ${\vec
\mu}^{(t)}$ is evolving over time according to a linear Gaussian model.
For simplicity, we assume that the $\Sigma^{\D(t)}$ which captures
time-specific topic correlations is independent across time.

It is noteworthy that unlike a standard SSM of which the latent state
would emit a single output (i.e., an observation or a measurement) at
each time point, the dMMSB model outlined above generates $N$
emissions each time, one corresponding to the (pre-transformed)
mixed-membership vector $\gammai^{\D(t)}$ of each vertex. To
directly apply the Kalman filter and Rauch--Tung--Striebel smoother
for posterior inference and parameter estimation under dMMSB, we
introduce an intermediate random variable $\vec{Y}^{(t)} = \frac{1}{N}
\sum_i
\gammai^{\D(t)}$; it is easy to see that $\vec{Y}^{(t)}$ follows a standard
SSM reparameterized from the original dMMSB: 
%
\begin{equation}
\vec{Y}^{(t)}  \sim\operatorname{Normal} \biggl(\vec{\mu}^{(t)}, \frac{\Sigma^{(t)}}{N} \biggr) \qquad t=1,\ldots,T.
\end{equation}

In principle, we can use the above membership evolution model to
capture not only membership correlation within and between vertices
at a specific time [as did in~\citet{bleiCTM}], but also dynamic
coupling (i.e., co-evolution) of membership proportions via
covariance matrix $\Phi$. In the simplest scenario, when $\Av=I$ and
$\Phi=\sigma I$, this model reduces to a random walk in the
membership-mixing space. Since in most realistic temporal series of
networks both the role-compatibility functions between vertices and the
semantic representations of membership-mixing are unlikely to be invariant
over time, we expect that even a random walk mixed-membership
evolution model can provide a better fit of the data than a static
model that ignores the time stamps of all networks.

\section{Variational inference}\label{sec:VI}

Due to difficulties in marginalization over the super-exponential state
space of latent variables $\vec{\zv}$ and $\vec{\pi}$, even the basic
MMSB model based on a Dirichlet prior over the role vectors $\vec{\pi}$
is intractable~[\citeauthor{xingLinkKDD05} (\citeyear{xingLinkKDD05,AiroBleiFienXing2008})]. With
the additional difficulty in integration of $\vec{\pi}$ under a
logistic normal prior where a closed-form solution is unavailable,
exact posterior inference of the latent variables of interest and
direct EM estimation of the model parameters is infeasible. In this
section we present a Laplace variational approximation scheme based on
the generalized means field (GMF) theorem~[\citet{xingGMF}] to infer the
latent variables and estimate the model parameters. This scheme
requires one additional approximating step on top of the variational
approximation developed in~\citet{AiroBleiFienXing2008}, but we will
show empirically in Section~\ref{sec:5.1} that this step does not
introduce much additional error. The GMF approach is modular, that is,
we can approximate the joint posterior
$p (\{\vec{\zv}^{\D(t)}, \vec{\pi}^{\D(t)}, \vec{\mu}^{\D(t)},
B^{\D(t)}\}_{\D t=1}^{\D T}|\Theta, \{G^{\D(t)}\}_{\D t=1}^{\D T}
)$, where $\Theta$ denotes the model parameters, by a factored
approximate distribution:
\begin{eqnarray}
&&q \bigl(\bigl\{\vec{\zv}^{\D(t)}, \vec{\pi}^{\D(t)}, \vec{\mu}^{\D(t)},
B^{\D(t)}\bigr\}_{\D t=1}^{\D T} \bigr)\nonumber
\\[-8pt]\\[-8pt]
&&\qquad= q_1 \bigl(\bigl\{\vec{\zv}^{\D(t)}, \vec{\pi}^{\D(t)}\bigr\}_{\D t=1}^{\D T} \bigr)
q_2 \bigl(\bigl\{\vec{\mu}^{\D(t)}\bigr\}_{\D t=1}^{\D T} \bigr)
q_3 \bigl(\bigl\{B^{\D(t)}\bigr\}_{\D t=1}^{\D T} \bigr),\nonumber
\end{eqnarray}
where $q_1(\cdot)$ can be shown to be the marginal distribution of $\{\vec
{\zv}^{\D(t)}, \vec{\pi}^{\D(t)}\}_{\D t=1}^{\D T}$ under a
reparameterized LNMMSB, and $q_2(\cdot)$ and $q_3(\cdot)$ are SSMs over $\{\vec
{\mu}^{\D(t)}\}_{\D t=1}^{\D T}$ and $\{B^{\D(t)}\}_{\D t=1}^{\D T}$,
respectively, with emissions related to expectation of $\{\vec{\zv}^{\D
(t)}, \vec{\pi}^{\D(t)}\}_{\D t=1}^{\D T}$ under $q_1(\cdot)$. This can be
shown by minimizing the Kulback--Leibler divergence between $q(\cdot)$ and
$p(\cdot)$ over arbitrary choices of $q_1(\cdot)$, $q_2(\cdot)$, and $q_3(\cdot)$, as
proven in~\citet{xingGMF}. The computation of the variational parameters
of each of these approximate marginals leads to a coupling of all the
marginals, as apparent in the descriptions in the subsequent
subsections. But once the variational parameters are solved, inference
on any latent variable of interest under the joint distribution $p(\cdot)$,
which is intractable, can be approximated by a much simpler inference
on the same variable in one of the $q_i(\cdot)$ marginals that contains the
variables of interest.
Below we briefly outline solutions to each of these marginals of subset
of variables, which exactly correspond to the three building blocks of
the dMMSB model outlined in Section~\ref{sec:dMMSB}. [Since $\mu^{\D
(t)}$ and $B^{\D(t)}$ both follow a standard SSM, for simplicity, we
only show the solution to $q_2(\cdot)$ over $\mu^{\D(t)}$, and treat $B^{\D
(t)}$ as an unknown invariant constant to be estimated.]

\subsection{Variational approximation to logistic-normal MMSB}\label{sec:inferlnmmsb}

For a static MMSB, the inference problem is to estimate the
role-vectors given model
parameters and observations. That is, model parameters $\vec\mu$,
$\Sigma$, and $B$ are assumed to be known besides the observed
variables $\Ev$, and we want to compute estimates of the role vectors
$\gammadot$ along with role indicators $\zdotTodot$ and $\zdotFromdot$.
(Under dMMSB, $\vec\mu$ is in fact unknown, but we will discuss shortly
how to estimate it outside of the MMSB inference detailed below.)

Under the LNMMSB, ignoring time and vertex indices, the marginal
posterior of latent variables $\vec{\gamma}$ (the pretransformed
$\vec{\pi}$) and $\vec{\zv}$ is
%
\begin{eqnarray}
\hspace*{15pt}&&p(\gammadot,\zdotTodot, \zdotFromdot | \vec\mu, \Sigma, B, \Ev)\nonumber
\\[-8pt]\\[-8pt]
&&\qquad\propto\prod_i p(\gammai | \vec\mu, \Sigma)
\prod_{i,j} p(\zitj, \zjfi | \gammai, \gammaj)p(\eij | \zitj,
\zjfi, B).\nonumber
\end{eqnarray}

Marginalization over all but one hidden variable to predict, say,
$\vec{\gamma}_i$, is intractable under the above model. Based on the GMF
theory, we approximate $p(\gammadot,\zdotTodot, \zdotFromdot | \vec\mu, \Sigma, B, \Ev)$
with a product of simpler marginals
$q(\cdot) =q_{\gamma}(\cdot)q_z(\cdot)$, each on a cluster of latent variable subsets,
that is, $\{\gammai\}$ and $\{\zitj,\break\zjfi\}$. \citet{xingGMF} proved
that under GMF approximation, the optimal solution, $q(\cdot)$, of each
marginal over the cluster of variables is isomorphic to the true
conditional distribution of the cluster given its \textit{expected Markov
Blanket}. That is,
%
\begin{eqnarray}
q_\gamma(\gammai)
& = & p(\gammai|\vec\mu, \Sigma, \Estz{\zitodot},
\Estz{\zifromdot}),\\
q_z(\zitj, \zjfi) & = & p(\zitj, \zjfi|\eij, B, \Estgamma{\gammai},
\Estgamma{\gammaj}).
\end{eqnarray}

These equations define a fixed point for $q_{\gamma}$ and $q_z$. The
optimal marginal distribution of the variables in one cluster is
updated when we fix the marginal of all the other variables, in turn.
The update continues until the change is neglectable.


The update formula for a cluster marginal of ($\zitj$, $\zjfi$) is
straightforward. It follows a multinomial distribution with $K\times K$
possible outcomes:\footnote{The $K\times K$ components are flatted into
a one-dimension vector.}
\begin{eqnarray}\label{eq:q_z}
\hspace*{25pt}q_z(\zitj, \zjfi)
&\propto& p(\zitj|\Estgamma\gammai)  p(\zjfi | \Estgamma
\gammaj)  p(\eij|\zitj, \zjfi, B)\nonumber \\[-8pt]\\[-8pt]
&\sim&\operatorname{Multinomial}(\vec\delta_{ij}), \nonumber
\end{eqnarray}
where $\delta_{ij(u,v)} \equiv\frac{1}{C} \exp( \Estgamma{\gamma_{i,u}}
+ \Estgamma{\gamma_{j,v}})  \Buv^{\eij}  (1-\Buv)^{1-\eij}$,
and $C$ is the normalization function to keep $\sum_{(u,v)}\delta_{ij(u,v)}=1$.
Furthermore, the expectation of $z$'s according to the multinomial
distribution are
\begin{eqnarray}\label{eq:z1}
 \Ezitju&=& \frac{\sum_v \delta_{ij(u,v)}}{\sum_{u,v}
\delta_{ij(u,v)}}=\sum_v \delta_{ij(u,v)},\nonumber
\\[-8pt]\\[-8pt]
\Ezjfiv&=& \frac{\sum_u \delta_{ij(u,v)}}{\sum_{u,v} \delta_{ij(u,v)}}=\sum_u \delta_{ij(u,v)}.\nonumber
\end{eqnarray}

The update formula for $\gammai$ can be derived similarly, but some
further approximation is applied. First,
\begin{eqnarray}\label{eq:q_gamma}
q_\gamma(\gammai)
& \propto & p(\gammai|\vec\mu, \Sigma)  p(\Estz\zitodot, \Estz
\zifromdot|\gammai) \nonumber\\[-8pt]\\[-8pt]
& = & \mathcal{N}(\gammai; \vec\mu, \Sigma)  \exp\bigl({\Estz{\vec{m}_i}^T}
\vec{\gamma_i} - (2N-2)  C(\vec{\gamma_i})\bigr), \nonumber
\end{eqnarray}
where $m_{ik} = \sum_{j\neq i} (\zitjs{k} + \ziFromjSubk)$, $\Estz
{m_{ik}} = \sum_{j\neq i}^N (\Estz{\zitjs{k}} + \Estz{\ziFromjSubk})$,
and $C(\vec{\gamma_i}) = \log(\sum_{k=1}^K \exp\{\gamma_{i,k}\} )$.
The presence of the normalization constant $C(\gammai)$ makes $q_\gamma
$ unintegrable in the closed-form.
Therefore, we apply a Laplace approximation to $C(\gammai)$ based on a
second-order Taylor expansion around $\hat{\gamma_i}$~[\citet{xingLoNTAM}],
such that $q_\gamma(\gammai)$ becomes a reparameterized multivariate normal
distribution $\mathcal{N}(\tildegamma_i, \tildeSigma_i)$ (see
Appendix~\ref{app:TaylorApprox} for details).
In order to get a good approximation, the
point of expansion, $\hat{\gamma_i}$, should be set as close to the
query point as possible. Therefore, we set it to be the $\tildegamma_i$
obtained from the previous iteration, that is, $\hatgamma_i^{r+1}
= \tildegamma_i^{r}$
where $r$ denotes the iteration number.

The inference algorithm iterates between equation~(\ref{eq:q_z}) and equation~(\ref{eq:q_gamma}) until convergence when the relative change of
log-likelihood is less than $10^{-6}$ in absolute value. The procedure
is repeated multiple times with random initialization for $\tildegamma
_i$. The result having the best likelihood is picked as the solution.

\subsection{Parameter estimation for logistic-normal MMSB}
\label{sec:inferlnmmsb1}

The model parameters $\vec\mu$, $\Sigma$, and $B$ have to be estimated
from data $\Ev\equiv\{e_{ij}\}$. In the simplest case, where time
evolution of $\vec\mu$ and $B$ is ignored, these can be done via a
straightforward EM-style procedure.

In the E-step, we use the inference algorithm from~Section~\ref{sec:inferlnmmsb} to compute
the posterior distribution and expectation of the latent variables by
fixing the current parameters.
In the M-step, we re-estimate the parameters by maximizing the
log-likelihood of the data using the posteriors obtained from the
E-step. Under a \mbox{LNMMSB}, exact computation of the log-likelihood is
intractable, hence, we use an approximation method known as variational
EM. We obtain the following update formulas for variational EM~[\citet{zoubin1}]
(see Appendix~\ref{app:sLearn} for an illustration of the derivation of the update for
$B$):
\begin{eqnarray}\label{est:mmsb}
\hatBkl&=& \frac{\sum_{i,j} \eij\delta_{ij(k,l)}}{\sum_{i,j} \delta_{ij(k,l)}},\qquad
\hat{\mu} = \frac{1}{N} \sum_i \tildegamma_i,\nonumber
\\[-8pt]\\[-8pt]
\hat{\Sigma} &=& \frac{1}{N} \sum_i \tildeSigma_i + \operatorname{Cov}(\tildegamma_{1:N}).\nonumber
\end{eqnarray}
%

The procedure for the learning can be summarized below.
%

Learning for logistic-normal MMSB:
\begin{enumerate}
\item initialize $B\sim\mathcal{U}[0,1]$, $\vec\mu~\sim\mathcal{N}(0,I)$, $\Sigma=10I$
\item while not converged (Outer Loop)
\begin{enumerate}[2.1.]
\item[2.1.]  Initialize $q(\gammai)$
\item[2.2.] while not converged and \#iteration $\le$ threshold (Inner Loop)
\begin{enumerate}[2.2.1.]
\item[2.2.1.] update $q(\zitj,\zjfi)\sim\operatorname{Multinomial}(\vec\delta_{ij})$
\item[2.2.2.] update $q(\gammai)\sim\mathcal{N}(\tildegamma_i, \tildeSigma_i)$
\item[2.2.3.] update $B$
\end{enumerate}
\item[2.3.] update $\vec\mu,\Sigma$
\end{enumerate}
\end{enumerate}
%

The convergence criterion is the same as in inference.
%
It is worth noting that the update of role-compatibility matrix $B$ is
in the inner loop, which means that it is updated as frequently as
mixed membership vectors $\gammai$. This makes sense because the
role-compatibility matrix and mixed membership vectors are closely coupled.\looseness=1

\subsection{Variational approximation to dMMSB}
\label{sec:infer_dmmsb}

When $\vec\mu$ is time-evolving as in dMMSB, two aspects in the
algorithms described in Sections~\ref{sec:inferlnmmsb} and~\ref{sec:inferlnmmsb1} need to be treated differently. First, unlike in
equation (\ref{est:mmsb}), estimation of $\vec{\mu}^{\D(t)}$ now must
be done under an SSM, with $\{\tildegamma^{\D(t)}_i\}$ as the
emissions at every time point. Second, according to the GMF theorem,
the $\mu$ that appeared in all equations in Section~\ref{sec:inferlnmmsb} must now be replaced by the posterior mean of $\vec
{\mu}^{\D(t)}$ under this SSM. Below we first summarize the algorithm
for dMMSB, followed by details of the update steps based on the Kalman
Filter (KF) and the Rauch--Tung--Striebel (RTS) smoother algorithms.\\

%

Inference for dMMSB:
\begin{enumerate}
\item initialize all $\vec{\mu}^{\D(t)}$
\item while not converged
\begin{enumerate}[2.1.]
\item[2.1.] for each $t$
\begin{enumerate}[2.1.1.]
\item[2.1.1.] call the inference algorithm for MMSB on network ${E}^{(t)}$ in Section~\ref{sec:inferlnmmsb}
      (by passing to it all current estimate of $\vec{\mu}^{\D(t)}$),
      and return the GMF approximation $\tildegamma_i^{(t)},\tildeSigma_i^{(t)}$
\item[2.1.2.] update the observations, $\vec{Y}^{(t)} = \sum_i \tildegamma_i^{(t)}/N$
\end{enumerate}
\item[2.2.] RTS smoother update $\vec\mu^{(t)} = \KFx_{t|T}$ based on $\{\vec{Y}^{(t)}\}_{\D t=1}^{\D T}$
\end{enumerate}
\end{enumerate}
%

Given all model parameters and all the emissions (the current estimate
of the mixed membership vectors $\{\tildegamma^{\D(t)}_i\}$ of all
vertices returned by the logistic-normal MMSB at each time point),
posterior inference of the hidden states $\vec{\mu}^{\D(t)}$ can be
solved according to the following KF and RTS procedure. The major
update steps in the Kalman Filter are as follows:
%
\begin{eqnarray}
\KFx_{t+1|t} & =& A \KFx_{t|t} = \KFx_{t|t} \nonumber,\\
\KFP_{t+1|t} & =& AP_{t|t}A^T+\Phi= \KFP_{t|t}+\Phi\nonumber,\\
K_{t+1} & =& \KFP_{t+1|t}(\KFP_{t+1|t}+\Sigma_{t+1} / N)^{-1} \nonumber,\\
\KFx_{t+1|t+1} & =& \KFx_{t+1|t} + K_{t+1}(\vec{Y}_{t+1}-\KFx_{t+1|t}) ,\\
\KFP_{t+1|t+1} & =& \KFP_{t+1|t} - K_{t+1} \KFP_{t+1|t},
\end{eqnarray}
where $\KFx_{t|s} \equiv\mathbb{E}(\vec\mu^{(t)}  | \vec{Y}_1,\ldots
,\vec{Y}_s)$
and $\KFP_{t|s} \equiv\operatorname{Var}(\vec\mu^{(t)}  | \vec{Y}_1,\ldots,\vec
{Y}_s)$. And the major update steps in the Rauch--Tung--Striebel
smoother are as follows:
%
\begin{eqnarray}
L_t & = & \KFP_{t|t}A^TP_{t+1|t}^{-1} = \KFP_{t|t}\KFP_{t+1|t}^{-1}
\nonumber,\\
\KFx_{t|T} & = & \KFx_{t|t} + L_t (\KFx_{t+1|T}-\KFx_{t+1|t}) ,\\
\KFP_{t|T} & = & \KFP_{t|t} + L_t (\KFP_{t+1|T}-\KFP_{t+1|t}) L_t^T.
\end{eqnarray}


\subsection{Parameter estimation for dMMSB}


We again use the variational EM algorithm. The E-step uses the dMMSB
inference algorithm in Section~\ref{sec:infer_dmmsb} for computing
sufficient statistics $\KFx_{t|T}, \forall t$, and the logistic normal
MMSB\vspace*{-1pt} inference algorithm in Section~\ref{sec:inferlnmmsb1} for
computing all sufficient statistics $\delta_{ij(k,l)}^{(t)}$. In the\vspace*{1pt}
M-step, model parameters are updated by maximizing the log-likelihood
obtained from the E-step. From this on, we simplify the linear
transition model posed on matrix $B$ and assume that it is constant. We
derive the following updates for the model parameters $B,\nu,\Phi,\Sigma
^{(t)}$ (see Appendix~\ref{app:tLearn} for some details):
%
\begin{eqnarray}\label{eq:learnB}
\hspace*{30pt} \hatBkl&=& \frac{\sum_t\sum_{i,j} \eij^{(t)}\delta_{ij(k,l)}^{(t)}}{\sum_t\sum_{i,j} \delta_{ij,(k,l)}^{(t)}}, \\
\label{eq:learnQ}
 \hat\Phi &=& \frac{1}{T-1}  \Biggl(\sum_{t=1}^{T-1}
(\KFx_{t+1|T}-\KFx_{t|T})(\KFx_{t+1|T}-\KFx_{t|T})^{\tt T} + \sum_{t=1}^{T-1} L_t\KFP_{t+1|T}L_t^T  \Biggr), \\
\label{eq:learnR}
 \hat\Sigma^{(t)} &=& \frac{1}{N}
\biggl( \sum_i \bigl(\KFx_{t|T}-\tildegamma_i^{(t)}\bigr)\bigl(\KFx_{t|T}-\tildegamma_i^{(t)}\bigr)^{\tt T} +
\sum_i \tildeSigma_i^{(t)}  \biggr), \\
 \hat\nu &=& \KFx_{1|T}.
\end{eqnarray}

The algorithm can be summarized below.

%

Learning for dMMSB:
\begin{enumerate}
\item initialize $B\sim\mathcal{U}[0,1]$, $\nu\sim\mathcal{N}(0,I)$, $\vec
\mu^{(t)}=\nu$, $\Phi=10I$, $\Sigma^{(t)}=10I$
\item  while not converged
\begin{enumerate}[2.1.]
\item[2.1.]  initialize all $q(\gammai^{(t)})$
\item[2.2.]  while not converged
\begin{enumerate}[2.2.1.]
\item[2.2.1.]  for each $t$
\begin{enumerate}[2.2.1.2.]
\item[2.2.1.1.]  update $q(\zitj,\zjfi)\sim\operatorname{Multinomial}(\vec\delta_{ij})$
\item[2.2.1.2.] update $q(\gammai)\sim\mathcal{N}(\tildegamma_i, \tildeSigma_i)$
\end{enumerate}
\item[2.2.2.] update $B$
\end{enumerate}
\item[2.3.] RTS smoother update, $\vec\mu^{(t)} = \KFx_{t|T}$ based on $\{\vec
{Y}^{(t)}\}_{\D t=1}^{\D T}$
\item[2.4.] update $\nu,\Phi,\Sigma^{(t)}$
\end{enumerate}
\end{enumerate}

Notice that in the above algorithm, the variational cluster marginals
$q(\zitj,\zjfi)$, $q(\gammai)$, and $q(\vec\mu^{(1)}, \ldots, \vec\mu
^{(T)})$ each depend on variational parameters defined by other cluster
marginals. Thus, overall the algorithm is essentially a fixed-point
iteration that will converge to a local optimum. We use multiple random
restarts to obtain a near global optimum.

%



\section{Experiments}\label{sec5}

In this section we validate the inference algorithms presented in
Section~\ref{sec:VI} on synthetic networks and demonstrate the
advantages of the dMMSB model on the well-known Sampson's monk network.
Then we apply dMMSB to two large-scale real world data sets.

\subsection{Synthetic networks}
We first evaluate the logistic normal MMSB described in Section~\ref{sec:sMMSB} in comparison with the earlier Dirichlet MMSB proposed
by~\citet{AiroBleiFienXing2008}, and then with the dMMSB model
described in Section~\ref{sec:dMMSB}. We investigate their differences
in three major aspects: (i) Is the Laplace variational inference
algorithm adequate for accurately estimating the mixed membership
vectors? (ii) For a static network, does LNMMSB provide a better
fit to the data when different roles are correlated? And (iii)
for dynamic networks, does dMMSB provide a better fit to the data?

\subsubsection{Inference accuracy}
\label{sec:5.1}
\begin{figure}[b]

\includegraphics{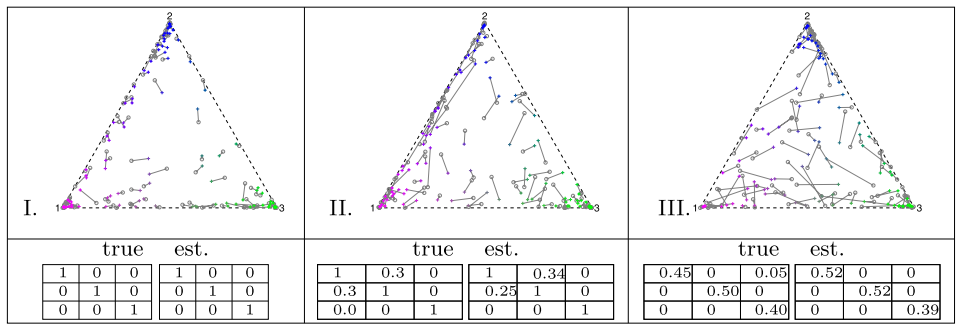}

\caption{Results of inference and learning with LNMMSB on
representative synthetic networks from scenario \textup{I} to \textup{III}. In the top
row, the figure in each cell displays the estimated role-vectors. They
are projected onto a simplex along with the ground truth: a circle
represents the position of a ground truth; a cross represents an
estimated position; and, each truth-estimation pair is linked by a grey
line. Note that we used different colors to denote actors from
different groups. In the bottom row, we display the the true and
estimated role-compatibility matrices. For all three cases, the
estimated role-compatibility matrices are close to the true matrices we
used to generate the synthetic networks.}
\label{fig:simu1}
\end{figure}

We generated three sets of synthetic networks, each of which has 100
individuals and 3 roles, using 3 different sets of role-vector priors
and role-compatibility matrices, to mimic different real-life
situations. Figure~\ref{fig:simu1} shows the estimation errors with
LNMMSB under the three scenarios. The results from the Dirichlet MMSB
are very close to that of LNMMSB and therefore are not shown here.

For synthetic network I, most actors have a single role and the
role-compatibility matrix is diagonal, which means that actors connect
mostly with other actors of the same role.
It can be seen that the mixed membership vectors are well recovered.
Most of the actors in the simplex are close to a corner, which
indicates that they have a dominating role. Some actors are not close
to a corner but close to an edge, which means that they have strong
memberships for two roles. The remaining actors lying near the center
of the simplex have mixed memberships for all three roles. In general,
the difficulty of recovering the mixed membership vector increases as
an actor possesses more roles.

In synthetic network II, the true mixed membership vector is
qualitatively similar to synthetic network I, but the
role-compatibility matrix contains off-diagonal entries. As a result,
an actor in network II is more likely to connect with actors of a
different role than network I. In this more difficult case, our model
still accurately estimates the role-compatibility matrix and the mixed
membership vectors.

In synthetic network III, we present a very difficult case where many
actors undertake noticeable mixed roles, and the within-role affinity
is very weak. Though a few actors near the center of the simplex endure
obvious discrepancy between the truth and the estimation, less than 10
percent of actors have more than 20 percent errors in their role
vectors. Furthermore, we can
see the group structure is still clearly retained.

\begin{figure}

\includegraphics{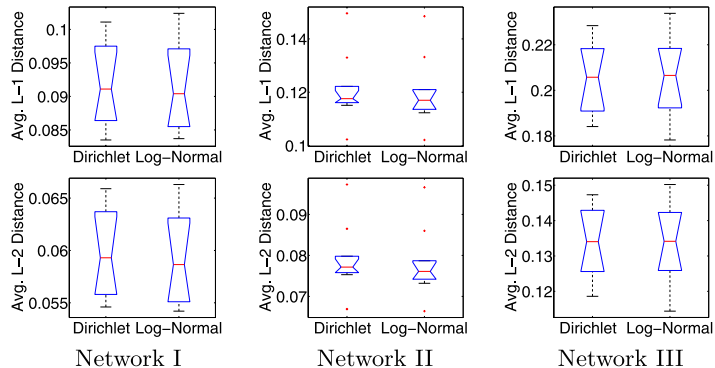}

\caption{The average distance in (top) L-1 and (bottom) L-2 between the
ground truth and the estimation of the mixed membership vectors in
networks that share parameter settings as simulation networks \textup{I, II} and
\textup{III} (from left to right).}\label{fig3} 
\end{figure}

Note that LNMMSB and Dirichlet MMSB employ different variational
schemes to approximate the posterior of the mixed membership vectors,
and the two models possess different modeling power to accommodate
correlations between different memberships. The combined effect could
lead to a difference in their accuracy of estimating the mixed
membership vectors of every vertex, although in practice we found such
difference hardly noticeable in the simplexial display given in
Figure~\ref{fig:simu1}. To provide a quantitative comparison between
the LNMMSB and the Dirichlet MMSB, we compute the average distance
between the ground truth and the estimated mixed membership vectors in
the aforementioned three settings. We used both the $\ell_1$ and the
$\ell_2$ distance as the metrics in our comparison, and the results are
shown in Figure~\ref{fig3}, where each type of network is
instantiated ten times to produce the error bar. We can see that the
LNMMSB performs slightly better for networks I and II (though no
significant difference is observed).

\subsubsection{Goodness of fit of LNMMSB}

To evaluate the fitness of the model to the data, we compute the
log-likelihood of fitting a type-II synthetic network generated in the
previous experiment, achieved by the model in question at convergence
of parameter estimation via the variational EM. Since no simple form of
the log-likelihood can be derived for both methods, the log-likelihoods
were obtained via importance sampling. The results for LNMMSB and
Dirichlet MMSB are listed in Table~\ref{tab:simu1}, showing that the
goodness of fit of the two models are comparable, with LNMMSB slightly
dominating over Dirichlet MMSB. As parallel evidence, the $\ell_2$ norm
distances
between the inferred mixed membership vectors and the ground truth are
also shown.

\begin{table}[b]
\tablewidth=7.2cm
\caption{Dirichlet vs. logistic normal prior for MMSB}\label{tab:simu1}
%
\begin{tabular*}{\tablewidth}{@{\extracolsep{\fill}}lcc@{}}
\hline
\textbf{Prior} & \textbf{Avg.} $\bolds{\ell_2}$ \textbf{distance} & \textbf{Log-likelihood} \\
\hline
Dirichlet           & 0.091 & $-$5755.8 \\
Logistic normal     & 0.092 & $-$5691.7 \\
\hline
\end{tabular*}
%
\end{table}

\subsubsection{Goodness of fit of dMMSB}
\begin{figure}

\includegraphics{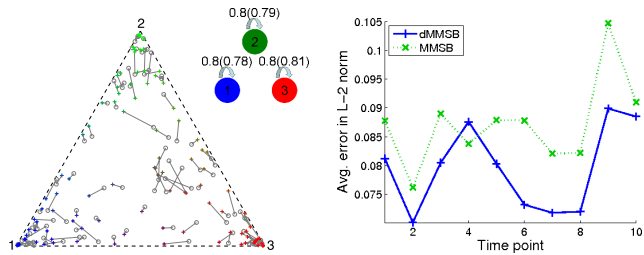}

\caption{\textup{Left}: The true mixed membership vectors (circle) and the
estimates by dMMSB (cross) at time point 6 visualized in a 2-simplex;
each truth-estimate pair is linked by a grey line.
\textup{Middle}: The learned role compatibility matrix, whose nonzero entries are shown by
arcs with values; values outside the brackets are the truths and the
values inside the brackets are estimates.
\textup{Right}: Average $\ell_2$ errors of mixed membership vectors for MMSB and dMMSB.}
\label{fig:simu2}
\end{figure}

To assess the fitness of the dMMSB, we generate dynamic networks
consisting of 10 time points. The number of actors remains 100 and the
number of roles remains 3. Furthermore, we generate the networks in
such a way that networks between adjacent time points show certain
degrees of similarity. As an illustration, the true role compatibility
matrix and the mixed membership vectors at time point 6 are displayed
in Figure~\ref{fig:simu2}.

In Figure~\ref{fig:simu2} (right), we compare dMMSB to an LNMMSB
learning a static network for each time point separately. We measure
the performance in terms of the average $\ell_2$ distance between the
estimates of the mixed membership vectors and their true values. It can
be seen that the error of dMMSB is lower than the error of MMSB in most
cases and about 10 percent lower on average. This suggests that dMMSB
can indeed integrate information across temporal domain and better
models the networks. More settings of model parameters have been tested
on both LNMMSB and dMMSB; they confirm that dMMSB is more effective in
modeling dynamic networks.

\subsection{Sampson's monk network: Emerging crisis in a cloister} Now
we illustrate the dMMSB model on a small-scale pedagogical example, the
Sampson network.
\citet{SampsonMonk} recorded the social interactions among a group of monks
while being a resident in a monastery. He collected a lot of
sociometric rankings on relations such as liking, esteem, praise, etc.
Toward the end of his study, a major conflict broke out and was
followed up by a mass departure of the members. The unique timing of
the study makes the data more interesting in the attempt to look for
omens of the separation.

We analyze the networks of liking relationship at three time points.
They contain 18 members (only junior monks). The networks are directed
rather than undirected, because one can like another while not vice versa.

We start with a static analysis on the network of time point 3, which
is the latest record before the crisis. Several researchers have also
studied the static network, including \citet{Breiger1975}, \citet
{White1976}, and \citet{AiroBleiFienXing2008}.

The network is fitted by our model with 1 to 5 roles.
The proper number of roles is selected by Bayesian Information
Criterion (BIC).

\begin{figure}

\includegraphics{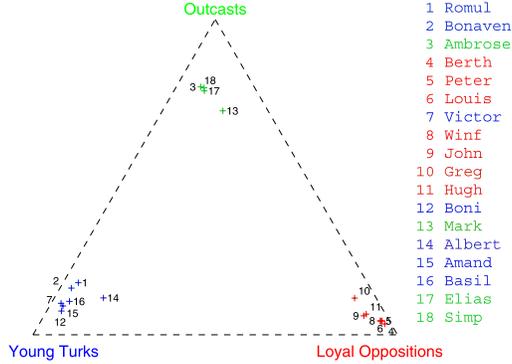}

\caption{Posterior mixed membership vectors of the monks projected in a
2-simplex by Log-Normal MMSB with 3 roles. Numbered points can be
mapped to monks' names using the legend on the right. Colors identify
the composition of mixed membership role-vectors.}\label{fig:monkLK_K3}
\end{figure}

Figure~\ref{fig:monkLK_K3} shows the posterior estimation of mixed
membership vectors of the monks in the monk liking networks by LNMMSB
with three roles. It clearly suggests three groups, each of which is
close to one vertex of the triangle. Using Sampson's labels, the three
groups correspond to the Young Turks (monks numbered 1, 2, 7, 12, 14,
15, 16), the Loyal Opposition (4, 5, 6, 9, 11) $+$ Waverers (8, 10), and
the Outcasts (3, 17, 18) $+$ Waverer (13). The result is consistent with
all previous works except for a controversial person, Mark (13). He is
known as an interstitial member of the monastery. \citet{Breiger1975}
placed him with the Loyal Opposition, whereas \citet{White1976} and
\citet{AiroBleiFienXing2008} placed him among the Outcasts.

\begin{figure}[b]

\includegraphics{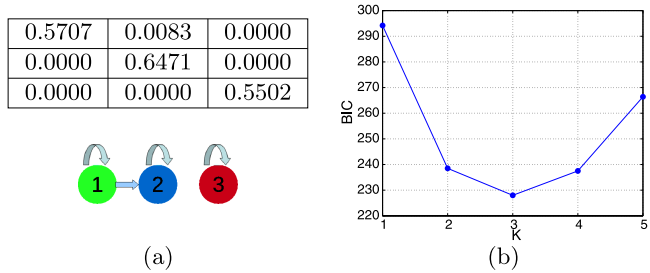}

\caption{\textup{(a)} The estimated role-compatibility matrix of the monk liking
networks by Log-Normal MMSB with 3 roles. \textup{(b)} The Bayesian Information
Criterion scores of the learning result of the monk liking network with
1 to 5 roles. The lower the better.}
\label{fig:monkLK_BIC}
\end{figure}

Figure~\ref{fig:monkLK_BIC}(a) demonstrates the estimated
role-compatibility matrix. It appears that the inter-group relation of
liking is strong, while the intra-group relation is absent. Together
with the fact that most of the individuals have an almost pure role, it
suggests that an explicit boundary exists between the groups, leaving
the later separation as no surprise.
Figure~\ref{fig:monkLK_BIC}(b) gives the BIC
scores. It suggests that the model with 3 roles is the best.

%

\begin{figure}

\includegraphics{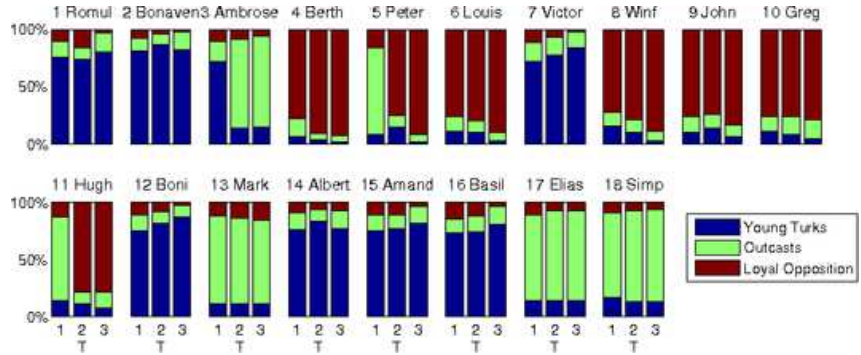}

\caption{The role-vectors learned in the dynamic network of liking
relationship between members in the Sampson Monastery. Each color
represents a role.}
\label{fig:TMonkLK_bar}
\end{figure}

The trajectories of the varying role-vectors over time inferred by
dMMSB with three roles are illustrated in Figure~\ref{fig:TMonkLK_bar}.
Several big changes in mixed membership vectors happened from time 1 to
time 2, and some minor fluctuation occurred between time 2 and time 3.
Overall, most persons were stable in the dominant role. If we only look
at time 3, which is the one we studied earlier in the static network
analysis, the results of mixed membership and grouping of the two
models are mostly consistent. Therefore, according to the discussion in
the static network analysis, the three roles in the dynamic model can
be roughly interpreted as Young Turks, Loyal Opposition, and Outcasts.

One of the persons whose dominant role changed is Ambrose (3). He later
became an Outcast. However, at time 1, he was connected with both Romul
(1) and Bonaven (2) in the Young Turks besides his connection with
Elias (17), an Outcast. It supports our result viewing him mainly as a
Young Turk at the time. The other two persons are Peter (5) and Hugh
(11). They were close to some Outcasts at time 1 but flipped to Loyal
Opposition at time 2, where they finally belonged to. It suggests that
the Outcast group whose member finally got expelled had not been
noticeably formed until after these big changes happened between time 1
and time~2.

From time 2 to time 3, it can be observed that the mixed membership
vectors were purifying, for instances, in monks numbered 1, 3--10, 12,
and 15--17. Bonaven (2) and Albert (14) were the exceptions, but they
did not change the general trend. The purifying process indicated that
the members of different groups were more and more isolated, which
finally led to the outbreak of a major conflict.

\subsection{Analysis of Enron email networks}

Now we study the Enron email communication networks. The email data was
processed by \citet{EnronUSC}. We further extract email senders and
recipients in order to build email networks. We have processed the data
such that numerous email aliases are properly corresponded to actual persons.

\begin{figure}

\includegraphics{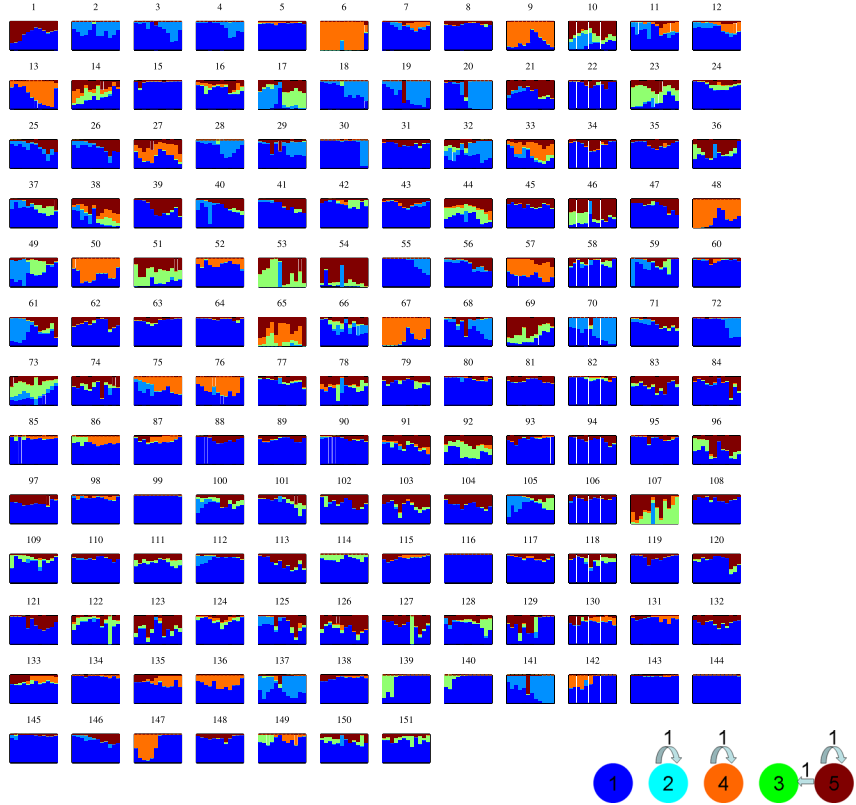}

\caption{Temporal changes of the mixed membership vectors for each
actor, and the visualization for role compatibility matrix.}
\label{fig:EnronB}
\end{figure}

There are 151 persons in the data set. We used emails from 2001, and
built an email network for each month, so the dynamic network has 12
time points.
We learn a dMMSB of 5 latent roles. The composition and trajectory of
roles of each recorded company employee and the role compatibility
matrix are depicted in Figure~\ref{fig:EnronB}.

It is observed that the first role (blue) stands for inactivity, that
is, the condition that a vertex is not interacting with any peers; this
is a necessary role to account for the intrinsic sparsity of the
network. The other roles are active. Actors with Role~2 (cyan), likely
representing lower-level employees, only send email to persons of the
same role, therefore, they form a clique. So is Role 4 (orange), which
leads to another clique. Persons \#6, 9, 48, 67, etc. mainly assume
this role, and they communicate with many others in the same role. They
appear to be normal employees according to available information and
the underlying meaning of the clique is yet to be discovered.

Role 5 (red) is within the functional composition of many people.
Persons in Role 5 send emails to persons with either Role 5 or Role 3
(green). They form a large clique, where Role 3 corresponds to
receivers and Role 5 to both senders and receivers. Role 3 might
reflect a certain aspect of senior management role that routinely
receives reports/instructions, while Role 5 might correspond to an
executive role that likes to issue orders to the managers and
communicate among themselves, or other level of positions that behave
somewhat similarly but possibly with opposite purpose, for example,
reporting to managers rather than dominating over them.

Of special interest are individuals that are frequently dominated by
multiple active roles (especially those falling into separate cliques),
because they have strong connection with different groups and may serve
important positions in the company. By scanning Figure~\ref{fig:EnronB}, actor \#65 and \#107 fit best to this category.
According to external sources, \textit{Mark Haedicke} (\#65) was the
Managing Director of the Legal Department and \textit{Louise Kitchen} (\#107) was the President of Enron Online, which supports the finding by
our method.

We also zoom into \textit{Kenneth Lay} (\#127), the Chairman and CEO of
Enron at the time. His role vector in August is abnormally dominated by
Role 3, which stands for a receiver. It is exactly the time when
Enron's financial flaws were first publicly disclosed by an analyst,
which might lead to a massive increase in enquiry emails from the
internal employees.

With respect to systematic changes in temporal space, the role vectors
of most actors are smooth over time. However, a few people experience a
large increase in the weight of the inactivity role in December (i.e.,
persons \#6, 13, 36, 67, 76). This is the time when Enron filed for
bankruptcy.

\begin{figure}

\includegraphics{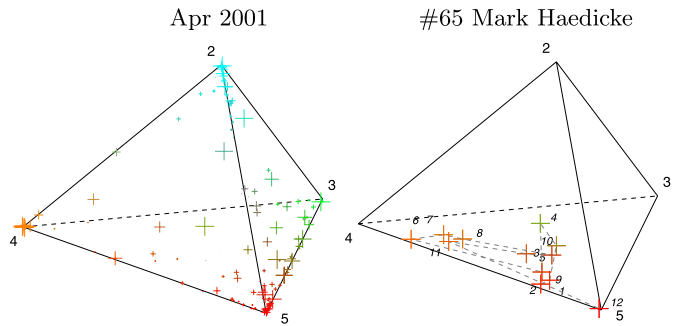}

\caption{\textup{Left}: Visualization of mixed membership vectors of
network actors in 3-simplex at one time point. Each vertex of the
tetrahedron corresponds to a role marked by its ID. A mixed membership
vector is represented by a cross whose location and color are the
weighted average of its active roles and whose size is proportional to
the sum of the weights from the active roles. \textup{Right}: We track the
trajectory of the mixed membership vector for an actor across time.
Numbers in italics show time stamps.}
\label{fig:EnronMix}
\end{figure}

We can also visualize the mixed membership vectors of the network
entities and track the trajectory of the mixed membership vector for an
individual as shown in Figure~\ref{fig:EnronMix}. They can help us
understand the network as a whole and how each individual evolves in
his or her role. Based on these examples, we believe dMMSB can provide
a useful visual portal for exploring the stories behind Enron.

\subsection{Analysis of evolving gene network as fruit fly aging}

In this section we study a sequence of gene correlation networks of the
fruit fly \textit{Drosophila melanogaster} estimated at various point of
its life cycle. It is known that over the developmental course of any
complex organism, there exist multiple underlying ``themes'' that
determine the functionalities of each gene and their relationships to
each other, and such themes are dynamical
and stochastic. As a result, the gene regulatory networks at each time point
are context-dependent and can undergo systematic rewiring, rather than
being invariant over time.
We expect the dMMSB model can capture such properties in the
time-evolving gene networks of
\textit{Drosophila melanogaster}.

However, experimentally uncovering the topology of the gene network at
multiple time points as the animal aging is beyond current technology.
Here we used the time-evolving networks of \textit{Drosophila
melanogaster} reverse-engineered by \citet{mladenAOAS} from a
genome-wide microarray time series of gene expressions using a novel
computational algorithm based on $\ell_1$ regularized kernel
reweighting regression, which is detailed in a companion paper that
also appears in this issue.
Altogether, 22 networks at different time points across various
developmental stages, namely, embryonic stage (1--10 time point),
larval stage (11--13 time point), pupal stage (14--19 time points), and
adult stages (20--22 time points), are analyzed. We focused on 588
genes that are known to be related to the developmental process based
on their gene ontologies.

\begin{figure}

\includegraphics{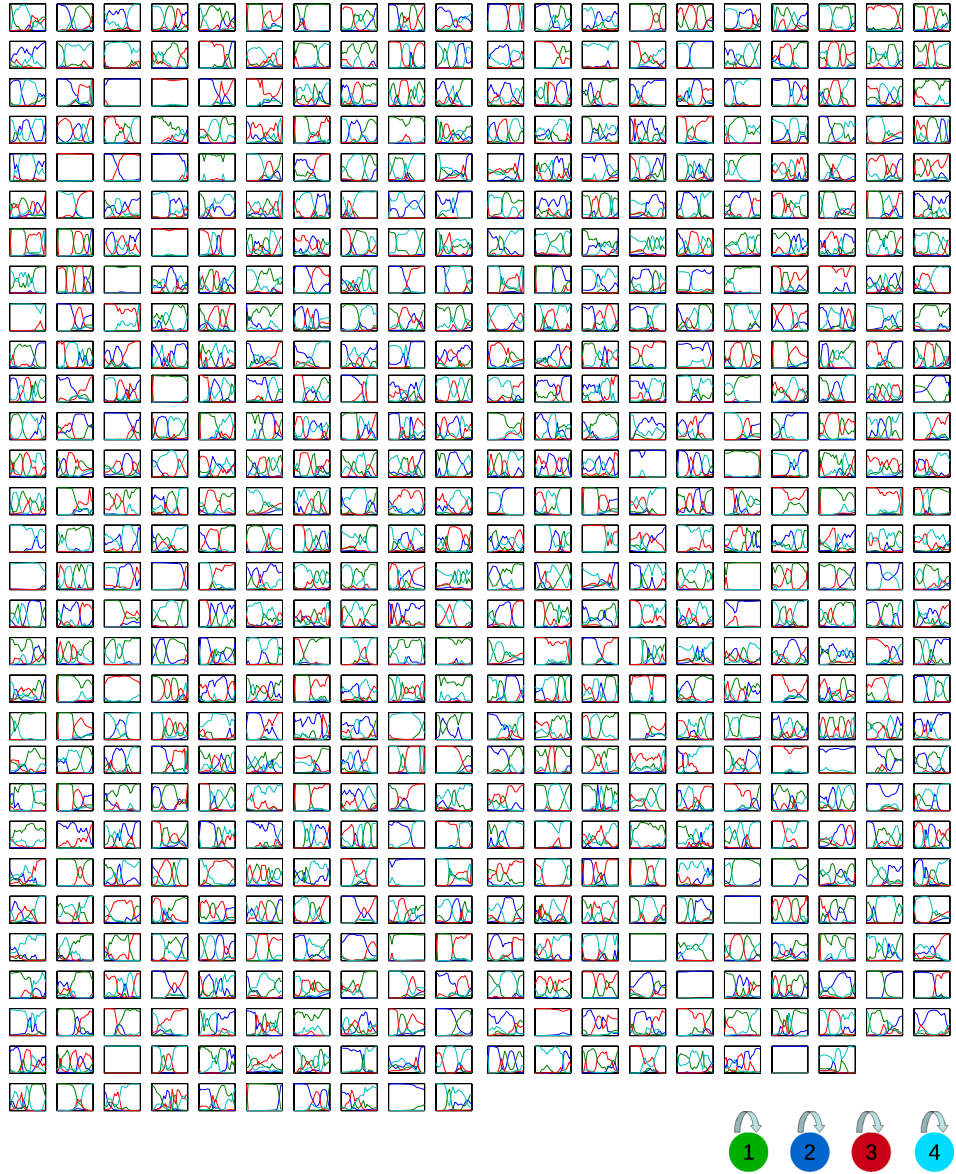}

\caption{Changes in mixed membership vectors of all genes, and the
visualization for role compatibility matrix. The $x$-axes of each subplot
is time, and the $y$-axes is the weight of role-component. Each color
stands for a role. }
\label{fig:DroVectors}
\end{figure}

We plotted the mixed membership vector over 4 roles for each gene as it
varies across the developmental cycle (Figure~\ref{fig:DroVectors}).
From the time courses of these mixed membership
vectors, we can see that many genes assume very different roles during
different stages of the development. In particular, we see that many genes
exhibit sharp transition in terms of their roles near the end of the
embryonic stage.
This is consistent with the underlying developmental requirement of
\textit{Drosophila} that the gene interaction networks need to undergo
a drastic reconfiguration to accommodate the new stage of larval
development. Somewhat surprisingly, we found when the number of roles
is set to four, the probability of interacting between different roles
is very small, as revealed by the visualization of the role
compatibility matrix (Figure~\ref{fig:DroVectors}, lower right). More
experiments are needed to examine whether this pattern is a true
property of the Drosophila gene interactions or an experimental
artifact (e.g., from accuracy of network reverse engineering, or from
the smallish number of roles we have chosen to fit the model, which
might be overly coarse, or from the quality of approximate inference in
a high-dimensional model).

%

\begin{figure}

\includegraphics{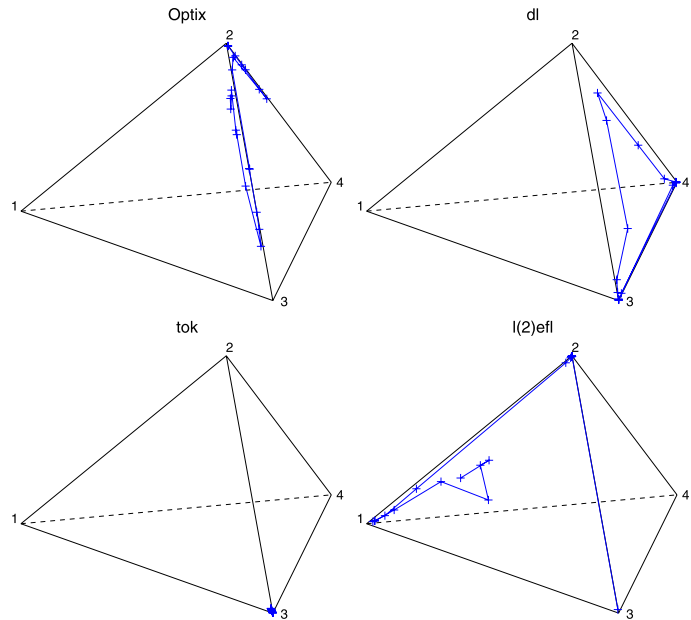}

\caption{The trajectories of mixed membership vectors of 4 genes
[Optix, dl, tok, l(2)efl].}
\label{fig:DroTraj}
\end{figure}

We selected four genes for further analysis, namely, Optix, dorsal
(dl), lethal (2) essential for life [l(2)efl], and tolkin (tok). These
four genes are among the highest degree nodes in the network produced
by averaging the dynamic networks over time. We want to see how their
roles evolve over time and, therefore, we plotted the trajectory of
their mixed membership vector in a 4-d simplex (Figure~\ref{fig:DroTraj}). We can see from the trajectory some of these genes
cover a wide area of the 4-d simplex. This is consistent with the roles
of gene Optix and dl as transcriptional factors that participate in
many different functions and regulate the expression of a wide range of
other genes. For instance, dl participates in a diverse range of
functions such as anterior/posterior pattern formation, dorsal/ventral
axis specification, immune response, gastrulation, heart development;
Optix participates in nervous system and compound eye development. In
contrast, gene tok and l(2)efl are not transcriptional factors and they
are currently only known for very limited functions: tok is related to
axon guidance and wing vein morphogenesis; l(2)efl is related to
embryonic and heart development. In our results, we found that, indeed,
the role-coordinates of tok are almost invariant, but the trajectory of
l(2)efl suggests that it may play more diverse roles than what is
currently known and deserves further experimental studies.

%
%

\begin{figure}

\includegraphics{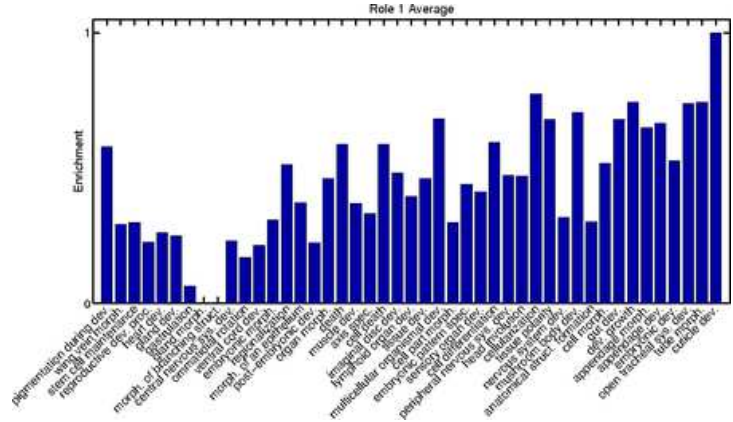}

\caption{Average gene ontology (GO) enrichment score for role 1. The
enrichment score for a given function is the number of genes labeled
as this function. Note that in the plot we have normalized the score
to a range between $[0,1]$, since we are mainly interested in
the relative count for each GO group.
Abbreviations appearing in
the figure are as follows: dev. for development, proc. for process,
morph. for
morphogenesis, and sys. for system.}
\label{fig:role1_avg}
\end{figure}

\begin{figure}[b]

\includegraphics{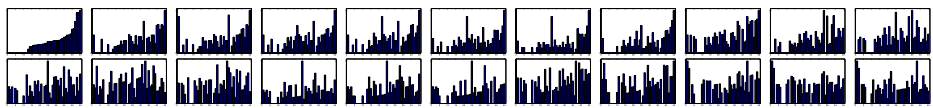}

\caption{Temporal evolution of gene ontology enrichment score for role
1. The time points are ordered from left to right, and
from top to bottom. The order of the gene ontology groups are
the same as in Figure~\protect\ref{fig:role1_avg}.}
\label{fig:role1_temp}
\end{figure}

We further used the mixture membership vectors as features to cluster
genes at each
time point into 4 clusters (each cluster corresponding to a particular
role-combination pattern),
and studied the gene functions in each role-combination across time. In
other words, we try to
provide a functional decomposition for each role obtained from the
dMMSB model
and investigate how these roles evolve over time.
In particular, we examined 45 ontological groups and computed the score
enrichment of these biological functions over random distribution in
each role cluster. Figures~\ref{fig:role1_avg} and~\ref{fig:role1_temp} demonstrate the results in cluster (i.e., role) 1. The
overall pattern that emerges from our results is that each role
consists of genes with a variety of functions, and the functional
composition of each role varies across time. However, the distributions
over these function groups are very different for different roles:
the most common functional groups for genes in role 1 are related to
multicellular
organismal development, cuticle development, and pigmentation during
development;
for the second role, the most common functional groups are gland
morphogenisis, heart development, gut development, and ommatidial
rotation; for the third role, they are stem cell maintenance, sensory
organ development, central nervous system development, lymphoid organ
development, and gland development; for the fourth role, gastrulation,
multicellular organismal development, gut development, stem cell
maintenance, and regionalization.


%

\section{Discussion}\label{sec6}
Unlike traditional descriptive methods for studying networks, which
focus on high-level ensemble properties such as degree distribution,
motif profile, path length, and node clustering, the dynamic mixed
membership stochastic blockmodel proposed in this paper offers an
effective way for unveiling detailed tomographical information of every
actor and relation in a dynamic social or biological network. This
methodology has several distinctive features in its structure and
implementation. First, the social or biological roles in the dMMSB
model are not independent
of each other and they can have their own internal dependency structures;
second, an actor in the network can be fractionally assigned to
multiple roles;
and third, the mixed membership of roles of each actor is allowed to vary
temporally. These features provide us extra expressive power to
better model networks with rich temporal phenomena.

In practice, this increased modeling power also provides better fit
to networks in reality. For instance, the interactions between genes
underlying the developmental course
of an organism are centered around multiple themes, such as wing
development and
muscle development, and these themes are tightly related to each other:
without the proper development of muscle structures, the development
and functionality of wings can not be fulfilled. As an organism moves
along its developmental cycle, the underlying themes can evolve and
change drastically.
For instance, during the embryonic stage of the \textit{Drosophila},
wing development is simply not present and other processes such as the
specification of anterior/posterior axis may be more dominant. Many
genes are
very versatile in terms of their roles and they differentially interact
with different genes
depending on the underlying developmental themes. Our model
is able to capture these various aspects of the dynamic gene
interaction networks,
and hence leads us a step further in understanding the biological processes.

In terms of the algorithm, a key ingredient to glue the three features
together is the logistic normal prior for the mixed membership
vector. This prior is superior to a Dirichlet prior in our context
since the
off-diagonal entries of the covariance matrix allow us to code the dependency
structure between roles, as clearly demonstrated in an earlier
work~[\citet{xingLoNTAM}]. Another advantage of the logistic normal
prior is that it can be readily coupled with a state-space model for
tracking the evolution of the roles. However, the drawback of the
logistic normal prior is
that it is not a conjugate prior to the multinomial distribution and, therefore,
additional approximation is needed during learning and inference. For
this purpose, we developed an efficient Laplace variational inference algorithm.

Our algorithm scales quadratically with the number of nodes in the
network, due to the necessity to infer the context-dependent role
indicator $Z$. It scales quadratically with the number of possible
roles, and linearly with the number of time steps, which are small
compared to the network size. The constant factor typically depends on
the stringency of the convergence test in the variational EM and the
number of random restarts to alleviate local optimum. In our current
implementation, we can handle a network with nodes $\sim$10$^3$ within a
day. We have been focusing on developing efficient algorithms that
enable dynamic tomographic analysis of ``meso-level'' networks, that is,
a network with thousands of nodes, rather than a ``mega'' network with
millions of nodes. We feel that this objective is appropriate because
for mega-networks, such as the blogsphere and the world wide web, it is
the ensemble behavior mentioned above that offers more important
information to an investigator who wants to do something with the
network, rather than individual nodal states. This change of focus with
the size of the system can also be seen in economics and game theory.

There are many dimensions where we can extend our current work. For instance,
the current model does not explicitly take hubs and cliques of the
networks into account, and
the state-space model does not enforce temporal smoothness directly
over the mixed
membership vector but only on its prior. Incorporating these elements
will be interesting future research.

\begin{appendix}

\section*{Appendix: Derivations}\label{appendix}

\subsection{Taylor approximation} \label{app:TaylorApprox}

We want to approximate $C(\gamma_i)$ by a second-order Taylor expansion.
For simplicity, we temporarily drop the subscript $i$ in this subsection.
The Taylor expansion of $C(\vec{\gamma})$ w.r.t. any point $\hatgamma$ is
%
\begin{equation}
C(\vec{\gamma}) \approx C(\hat{\gamma})+{\vec{g}}^T(\vec\gamma-\hatgamma)
+\tfrac{1}{2}(\vec\gamma-\hatgamma)^TH(\vec\gamma-\hatgamma),
\end{equation}
where $\vec{g}$ is the first derivative (a $K \times1$ vector), and
$H$ is the second derivative (a~$K \times K$ matrix). Only linear and
quadratic terms are left. Therefore, equation~(\ref{eq:q_gamma}) becomes
\begin{eqnarray*}
q_\gamma(\vec{\gamma})
& \propto&\mathcal{N}(\vec{\gamma}; \vec\mu, \Sigma)
\exp\bigl({\Estz{\vec{m}}^T }\vec{\gamma} - (2N-2)  C(\vec{\gamma})\bigr) \\
& \approx&\exp \bigl\{ -\tfrac{1}{2}(\vec{\gamma}-\vec{\mu})^T\Sigma
^{-1}(\vec{\gamma}-\vec{\mu})+{\vec r}^T\vec{\gamma}+\vec{\gamma
}^TS\gamma \bigr\},
\end{eqnarray*}
where ${\vec r}^T = \Estz{m}^T-(2N-2){\vec{g}}^T+(2N-2)\hatgamma^T H$
is a 1$\times K$ row vector and $S = -(N-1)H$ is a $K\times K$
symmetric matrix.

Letting  $x=\vec{\gamma}-\vec{\mu}$, the exponent becomes
\begin{eqnarray*}
&& -\tfrac{1}{2}(\vec{\gamma}-\vec{\mu})^T\Sigma^{-1}(\vec{\gamma}-\vec{\mu})+{\vec r}^T\vec{\gamma}+\vec{\gamma}^TS\gamma\\
&&\qquad=  -\tfrac{1}{2}x^T\Sigma^{-1}x+{\vec r}^T(x+\vec{\mu})+(x+\vec{\mu})^TS(x+\vec{\mu}) \\
&&\qquad=  -\tfrac{1}{2}x^T(\Sigma^{-1}-2S)x+({\vec r}^T+2{\vec{\mu}}^TS)x+C_1\\
&&\qquad\phantom{={}}(\mbox{and letting } \tildeSigma^{-1}=\Sigma^{-1}-2S,D={\vec r}^T+2{\vec{\mu}}^TS) \\
&&\qquad=  -\tfrac{1}{2}x^T\tildeSigma^{-1}x+Dx+C_1 \\
&&\qquad=  -\tfrac{1}{2}(x-\tildeSigma D^T)^T\tildeSigma^{-1}(x-\tildeSigma D^T)+C_2 \\
&&\qquad=  -\tfrac{1}{2}(\vec{\gamma}-\vec{\mu}-\tildeSigma D^T)^T\tildeSigma^{-1}(\vec{\gamma}-\vec{\mu}-\tildeSigma D^T)+C_2.
\end{eqnarray*}
Therefore, $\tildeSigma = (\Sigma^{-1}-2S)^{-1} =  (\Sigma^{-1}+(2N-2)H)^{-1}$
\begin{eqnarray*}
\tildegamma &=& \vec{\mu}+\tildeSigma D^T = \vec{\mu}+\tildeSigma(A^T+2S\vec{\mu})\\
 &=& \vec\mu+\tildeSigma \bigl({\langle\vec{m}_i\rangle}_{q_z}-(2N-2)\vec{g}+(2N-2)H\hatgamma_i-(2N-2)H\vec\mu \bigr),
\end{eqnarray*}
where the first and the second derivatives are
\begin{eqnarray*}
g(\hatgamma)_k & = &\frac{\exp\hatgamma_k}{\sum_k\exp\hatgamma_k},\label{eq:g_gamma}
\\
H(\hatgamma)_{kl} & = &\frac{\mathbb{I}(k=l)}{\sum_k\exp\hatgamma_k}-\frac{\exp\hatgamma_k \exp\hatgamma_l}{(\sum_k\exp\hatgamma_k)^2} \\
\end{eqnarray*}
or, in short,
\[
H  =  \operatorname{diag}(\vec{g})-\vec{g}\vec{g}^T.
\]

\subsection{Learning on logistic-normal MMSB} \label{app:sLearn}

The log-likelihood as a function of $B$ can be written as
%
\begin{eqnarray}
l(B) & = &\sum_{i,j}\log\sum_{k,l}  \bigl(\delta_{ij,(k,l)} \Bkl^{\eij}(1-\Bkl)^{(1-\eij)} \bigr) +C_0 \nonumber\\
& \ge&\sum_{i,j}\sum_{k,l} \delta_{ij,(k,l)} \log \bigl(\Bkl^{\eij}(1-\Bkl)^{(1-\eij)} \bigr) +C_0 \nonumber\\[-8pt]\\[-8pt]
& =& \sum_{i,j}\sum_{k,l} \delta_{ij,(k,l)}  \bigl(\eij\log\Bkl+ (1-\eij)\log(1-\Bkl)  \bigr) +C_0 \nonumber\\
& \equiv& l^*(B), \nonumber\\
\frac{\partial l^*(B)}{\partial\Bkl}
& = &\sum_{i,j}\sum_{k,l} \delta_{ij,(k,l)}  \biggl(\frac{\eij}{\Bkl} - \frac{1-\eij}{1-\Bkl}  \biggr),\nonumber\\
\hatBkl& =& \frac{\sum_{i,j} \eij\delta_{ij,(k,l)}}{\sum_{i,j} \delta_{ij,(k,l)}}.
\end{eqnarray}
Jensen's Inequality is applied in the derivation to get an
approximation (more specifically, a lower bound) to the log-likelihood
which has an analytical solution in finding the maximum point. Setting
the derivative to zero gives us an MLE estimator of $B$ based on approximation.

\subsection{Learning on dMMSB} \label{app:tLearn}

Again, we take an approximation of the log-likelihood, which is more tractable:
\begin{eqnarray}
l(B) & =& \sum_t\sum_{i,j}\log\sum_{k,l}  \bigl(\delta_{ij,(k,l)}^{(t)}
\Bkl^{\eij^{(t)}} (1-\Bkl)^{(1-\eij^{(t)})} \bigr) +C_0 \nonumber\\
&\ge&\sum_t\sum_{i,j}\sum_{k,l} \delta_{ij,(k,l)}^{(t)} \log \bigl(\Bkl^{\eij^{(t)}} (1-\Bkl)^{(1-\eij^{(t)})} \bigr) +C_0 \nonumber\\[-8pt]\\[-8pt]
& =& \sum_t\sum_{i,j}\sum_{k,l} \delta_{ij,(k,l)}^{(t)}  \bigl(\eij^{(t)}\log\Bkl+ \bigl(1-\eij^{(t)}\bigr)\log(1-\Bkl)  \bigr) +C_0 \nonumber\\
& \equiv& l^*(B).\nonumber
\end{eqnarray}
The update equation for $B$ is from maximizing the upper bound of the
log-likelihood:
\begin{eqnarray}
\frac{\partial l^*(B)}{\partial\Bkl}
& =& \sum_t\sum_{i,j}\sum_{k,l}\delta_{ij,(k,l)}^{(t)}  \biggl(\frac{\eij^{(t)}}{\Bkl} - \frac{1-\eij^{(t)}}{1-\Bkl}  \biggr), \nonumber\\
\hatBkl& =& \frac{\sum_t\sum_{i,j} \eij^{(t)}\delta_{ij,(k,l)}^{(t)}}{\sum_t\sum_{i,j} \delta_{ij,(k,l)}^{(t)}}.
\end{eqnarray}
\end{appendix}

\printaddresses

\end{document}